\documentclass{article} 
\usepackage[T1]{fontenc}
\usepackage{subcaption}
\usepackage[table]{xcolor}
\usepackage{colm2024_conference}
\usepackage{bbm}
\usepackage{microtype}
\usepackage{hyperref}
\usepackage{graphicx}
\usepackage{url}
\usepackage{cuted}
\usepackage[most]{tcolorbox}
\usepackage{booktabs}

\usepackage{amssymb}       
\usepackage{bbm}
\usepackage{pifont}        

\definecolor{darkblue}{rgb}{0, 0, 0.5}
\hypersetup{colorlinks=true, citecolor=darkblue, linkcolor=darkblue, urlcolor=darkblue}

\title{EfficientXLang:  Towards Improving Token Efficiency Through Cross-Lingual Reasoning}

\author{Sanchit Ahuja\thanks{Equal Contribution.}, Praneetha Vaddamanu*, Barun Patra \\
Microsoft Corporation \\
\texttt{\{sanchitahuja205@gmail.com, pvaddamanu@microsoft.com, bpatra@microsoft.com\}}
}

\colmfinalcopy 
\begin{document}

\maketitle

\begin{abstract}
Despite recent advances in Reasoning Language Models (RLMs), most research focuses solely on English, even though many models are pretrained on multilingual data. In this work, we investigate: \textit{Is English the most token-efficient language for reasoning?} We evaluate three open-source RLMs: \textsc{DeepSeek R1}, \textsc{Qwen 2.5}, and \textsc{Qwen 3}, across four math datasets and seven typologically diverse languages. We find that reasoning in non-English languages not only reduces token usage, but also preserves accuracy. These gains persist even after translating the reasoning traces into English, suggesting genuine shifts in reasoning behavior rather than surface-level linguistic effects. The extent of improvement, however, depends on the model’s multilingual strength. Our findings motivate a broader view of reasoning in language models, highlighting the potential of multilingual reasoning and the importance of strong multilingual foundations. The code for our work can be found \href{https://github.com/microsoft/EfficientXLang}{here}.
\end{abstract}
\section{Introduction}

Recent advancements have shown that increasing inference-time compute can significantly enhance the downstream performance of large language models. In particular, RLMs benefit from extended test-time computation through long chains-of-thought, resulting in improved reasoning capabilities and task performance \citep{deepseekai2025deepseekr1incentivizingreasoningcapability, openai2024openaio1card}. Despite this progress, the majority of prior work has focused on reasoning in English, implicitly assuming it as the default or optimal language for thought \citep{wendler2024llamas}. 

A key limitation of current RLMs lies in the number of ``thinking tokens'' they consume during inference. Since test-time compute scales with the number of tokens generated, longer reasoning traces directly impact latency and cost. This inefficiency is often exacerbated by the phenomenon of overthinking \citep{dangerofoverthinking}, where models produce unnecessarily verbose intermediate reasoning. To address this, several works have explored length-aware rewards in reinforcement learning setups to encourage brevity in reasoning traces \citep{aggarwal2025l1, sui2025stop}. However, these approaches typically require model retraining, which is computationally expensive and infeasible for many practitioners.

In this work, we take a different perspective on language and reasoning by asking a simple yet underexplored question: \textit{Is English the most efficient language for reasoning?} We explore whether reasoning in other languages can produce shorter but equally effective reasoning traces, potentially reducing inference-time compute. This question is timely for two reasons: (1) many reasoning language models (RLMs) are multilingual by design \citep{qwen3}; and (2) prior work suggests that languages vary in how clearly they express mathematical ideas, affecting reasoning performance \citep{han2001chinese}. Our key contributions are:

\setlength{\parindent}{5pt}

\textbf{-} We show that across a range of mathematical reasoning datasets, reasoning in alternative languages consistently results in fewer tokens, often matching the final English accuracy.

\textbf{-} We demonstrate that even when translated back to English, reasoning traces originally generated in other languages remain shorter than those natively generated in English. This suggests that the reduction in token count is not merely a function of tokenizer fertility or linguistic compactness, but reflects genuine differences in the characteristics of reasoning traces compared to their English counterparts.

\textbf{-} We provide quantitative evidence that English is not always the most effective language for reasoning, with traces in other languages consistently preferred for their conciseness, clarity, and structure.

\label{sec:analysis}
\begin{table*}[th]
\centering
\tiny

\subfloat[Performance relative to English (\texttt{TLP@4} and Tok. as \% change).]
{
\begin{tabular}{@{}l|ccc|ccc|ccc|ccc@{}}
\toprule
\textbf{Model} & \multicolumn{3}{c|}{\textbf{en}} & \multicolumn{3}{c|}{\textbf{zh}} & \multicolumn{3}{c|}{\textbf{ru}} & \multicolumn{3}{c}{\textbf{es}} \\
\cmidrule(lr){2-4} \cmidrule(lr){5-7} \cmidrule(lr){8-10} \cmidrule(lr){11-13}
& \texttt{TLP@4} & TLC & Tok. & $\Delta$\%TLP@4 & TLC & $\Delta$\%Tok & $\Delta$\%TLP@4 & TLC & $\Delta$\%Tok & $\Delta$\%TLP@4 & TLC & $\Delta$\%Tok \\
\midrule
\textsc{DeepSeek R1} & 0.87 & 1.000 & 3816 & 4.6\% & 1.000 & 21.1\% & 4.3\% & 1.000 & 14.1\% & -0.9\% & 0.985 & 29.9\% \\
\textsc{Qwen 2.5}     & 0.87 & 1.000 & 6249 & 3.8\% & 0.796 & 21.4\% & 23.3\% & 0.340 & 38.2\% & 82.0\% & 0.056 & 88.1\% \\
\textsc{Qwen 3}     & 0.89 & 0.998 & 6123 & 0.9\% & 0.987 & 27.7\% & -0.3\% & 0.996 & 35.8\% & 1.7\% & 0.583 & 73.0\% \\
\bottomrule
\end{tabular}
}
\hfill
\subfloat[Performance relative to English (\texttt{TLP@4} and Tok. as \% change).]
{
\begin{tabular}{@{}l@{}|ccc|ccc|ccc|ccc@{}}
\toprule
\textbf{Model} & \multicolumn{3}{c|}{\textbf{ar}} & \multicolumn{3}{c|}{\textbf{tr}} & \multicolumn{3}{c|}{\textbf{hi}} & \multicolumn{3}{c}{\textbf{ko}} \\
\cmidrule(lr){2-4} \cmidrule(lr){5-7} \cmidrule(lr){8-10} \cmidrule(lr){11-13}
& $\Delta$\%TLP@4 & TLC & $\Delta$\%Tok & $\Delta$\%TLP@4 & TLC & $\Delta$\%Tok & $\Delta$\%TLP@4 & TLC & $\Delta$\%Tok & $\Delta$\%TLP@4 & TLC & $\Delta$\%Tok \\
\midrule
\textsc{DeepSeek R1} & 7.4\% & 0.901 & 28.2\% & 8.3\% & 1.000 & 8.3\% & 8.0\% & 0.971 & 7.7\% & 9.5\% & 0.997 & 43.9\% \\
\textsc{Qwen 2.5}     & 57.2\% & 0.177 & 68.1\% & 29.3\% & 0.444 & 52.3\% & 58.3\% & 0.097 & 41.3\% & 81.4\% & 0.059 & 90.1\% \\
\textsc{Qwen 3}     & 15.7\% & 0.666 & 47.4\% & 18.5\% & 0.721 & 59.6\% & 16.3\% & 0.887 & 23.7\% & 17.1\% & 0.574 & 47.4\% \\
\bottomrule
\end{tabular}
}

\caption{Performance relative to English: \texttt{TLP@4} and mean token count are reported as percentage change from English; Target Language Consistency (TLC) is reported directly.}
\label{tab:language-performance-relative}
\end{table*}
\definecolor{green-1}{RGB}{213, 216, 123}
\definecolor{green-2}{RGB}{184, 209, 143}
\definecolor{green-3}{RGB}{152, 199, 134}
\definecolor{green-4}{RGB}{101, 189, 110}
\definecolor{yellow-1}{RGB}{255, 234, 132}
\definecolor{yellow-2}{RGB}{254, 217, 129}
\definecolor{orange-1}{RGB}{254, 206, 127}
\definecolor{orange-2}{RGB}{252, 173, 120}
\definecolor{red-1}{RGB}{247, 121, 121}

\begin{table*}[t]
\centering
\resizebox{\textwidth}{!}{%
\begin{tabular}{l|cc|cc|cc|cc|cc|cc|cc|cc}
\toprule
& \multicolumn{2}{c|}{\textbf{en}} & \multicolumn{2}{c|}{\textbf{ko}} & \multicolumn{2}{c|}{\textbf{es}} & \multicolumn{2}{c|}{\textbf{zh}} & \multicolumn{2}{c|}{\textbf{hi}} & \multicolumn{2}{c|}{\textbf{ru}} & \multicolumn{2}{c|}{\textbf{tr}} & \multicolumn{2}{c}{\textbf{ar}} \\
\cmidrule{2-17}
\textbf{Dataset} & \textbf{\texttt{TLP@4}} & \textbf{Tokens}
& $\Delta$\%TLP@4 & $\Delta$\%Tok
& $\Delta$\%TLP@4 & $\Delta$\%Tok
& $\Delta$\%TLP@4 & $\Delta$\%Tok
& $\Delta$\%TLP@4 & $\Delta$\%Tok
& $\Delta$\%TLP@4 & $\Delta$\%Tok
& $\Delta$\%TLP@4 & $\Delta$\%Tok
& $\Delta$\%TLP@4 & $\Delta$\%Tok \\
\midrule
\textbf{MATH500} & 0.77 & 1404.24
& \cellcolor{orange-2}0.00 & 37.80
& \cellcolor{orange-2}0.00 & 32.85
& \cellcolor{orange-2}0.00 & 18.41
& \cellcolor{orange-2}0.00 & 17.81
& \cellcolor{orange-2}0.00 & 36.96
& \cellcolor{orange-2}0.00 & 5.78
& \cellcolor{orange-2}0.00 & 26.98 \\
\textbf{AMC23} & 1.00 & 3571.1
& \cellcolor{orange-1}3.00 & 42.09
& \cellcolor{orange-2}0.00 & 25.20
& \cellcolor{orange-2}1.00 & 14.44
& \cellcolor{orange-2}2.00 & -0.83
& \cellcolor{orange-2}3.00 & 27.22
& \cellcolor{orange-1}4.00 & -12.29
& \cellcolor{orange-1}4.00 & 25.87 \\
\textbf{AIME2024} & 0.89 & 5637.27
& \cellcolor{yellow-1}12.36 & 42.40
& \cellcolor{orange-1}4.49 & 21.95
& \cellcolor{yellow-1}11.24 & 27.07
& \cellcolor{yellow-1}12.36 & -13.58
& \cellcolor{yellow-2}8.99 & 0.57
& \cellcolor{yellow-1}11.24 & -11.08
& \cellcolor{yellow-1}11.24 & 17.43 \\
\textbf{AIME2025} & 0.81 & 7979.17
& \cellcolor{green-2}24.69 & 47.78
& \cellcolor{orange-1}2.47 & 36.54
& \cellcolor{yellow-2}4.94 & 20.43
& \cellcolor{green-1}18.52 & 25.79
& \cellcolor{yellow-2}4.94 & 13.29
& \cellcolor{green-1}18.52 & 32.13
& \cellcolor{yellow-2}13.58 & 37.24 \\
\bottomrule
\end{tabular}%
}
\caption{Both TLP@4 and mean token count are reported as percentage changes from English. Color intensity ranges from orange to green/yellow, where brighter shades (green/yellow) indicate greater deviation from English performance, while duller shades like orange represent minimal or no deviation.}
\label{tab:tlp-delta_deepseek}
\end{table*}

\section{Experiments}
\label{sec:experiments}

To assess whether reasoning in other languages can enhance thinking token efficiency, we experiment with seven typologically diverse languages: Chinese, Russian, Spanish, Hindi, Arabic, Korean, and Turkish, along with English. We prompt the reasoning models in English, instructing them to carry out their reasoning in the target language while providing the final answer in English. This setup allows us to isolate the effect of language on the reasoning process itself. For inferencing, we used the SGLang framework \cite{NEURIPS2024_724be447}, which facilitates the extraction of intermediate reasoning traces.\footnote{The exact prompt we use can be found in Appendix \ref{sec:appendix}.} We now detail our experimental setup.

\setlength{\parindent}{0pt}


\textbf{Datasets}: We evaluate our hypothesis in a variety of math datasets, which we categorize into \textit{easy} and \textit{difficult} subsets based on the complexity of the problem and the typical performance of existing models. The \textit{difficult} datasets include AIME2024~\cite{aime_1983_2024} and AIME2025~\cite{aime2025_i, aime2025_ii}, which feature competition-level problems that require advanced reasoning. The \textit{easy} datasets include AMC23 \cite{aops2023amc12a, aops2023amc12b}, and MATH500 \cite{lightman2023lets} comprising of simpler problems.

\textbf{Models}: We explore three open-weight reasoning models for our experiments: \textsc{Deepseek R1} \cite{deepseekai2025deepseekr1incentivizingreasoningcapability}, QwQ (based on the 32B variant of \textsc{Qwen 2.5}) \citep{qwen2.5, qwq32b}, and \textsc{Qwen3} (Qwen3-235B-A22B), the latest model focused on reasoning of the Qwen family \cite{qwen3}.


\textbf{Evaluation}: Following the evaluation setup of \textsc{Deepseek R1}, we set the maximum generation length to 32,768 tokens. For response generation, we use a sampling temperature of 0.6 and a top-$p$ value of 0.95, generating $n = 16$ responses across all models and datasets.



\textbf{Metrics:} Different RLMs vary in their ability to follow prompt instructions for generating reasoning traces in the indicated target language. To better evaluate multilingual reasoning performance, we introduce two complimentary metrics: Target Language Consistency (TLC), and Target Language Pass @k (TLP@k) in conjunction with the commonly used Pass@k metric. We use $k=4$ in all our results.
\setlength{\parindent}{10pt}

\textbf{Target Language Consistency (TLC)}: measures how well models adhere to the target language constraint when prompted. Given a dataset of size $m$, with question-answer pairs $(q_i, y_i)$, and target language $j$, let $\mathcal{R}^{(k)}(i, j) = \{r_1, ..., r_k\}$ be the set of generated reasoning traces under the prompt to reason in language $j$. TLC for a sample is the fraction of traces in the correct language, and the overall TLC is the average across all samples. We use a LID model \cite{joulin2016bag} to detect the language of a reasoning trace.

\textbf{Target Language Pass @k (TLP@k)}: Analogous to the Pass@k metric, we propose TLP@k to measure the fraction of samples where at least one of the top-k reasoning traces both elicits the correct answer and is in the target language. Extending \citet{chen2021evaluatinglargelanguagemodels}, for measuring TLP@k, we use a low variance estimator $\frac{1}{m} \sum_{i=1}^{m}\left[1 - \frac{\binom{n - c}{k}}{\binom{n}{k}}\right]$, where $c$ is the number of reasoning traces that yield the correct answer and are in the correct target language among the $n$ generated traces. The final TLP@k is the average across all samples. Further details about the implementation of this estimator are provided in the Appendix \ref{app:unbiased_estimator}.

\section{Analysis}

This section explores a series of research questions aimed at validating our thesis.

\setlength{\parindent}{0pt}
\textbf{RQ1: Is English the optimal language for reasoning for LRMs ?} Table~\ref{tab:language-performance-relative} shows per-language model performance relative to English (absolute scores in Table~\ref{tab:language-performance}). Across models, high TLC correlates with smaller performance drops when reasoning in non-English languages—despite using significantly fewer tokens. In contrast, models with weaker multilingual foundations (e.g., \textsc{Qwen 2.5}) show larger performance gaps.

\begin{figure}[ht]
    \centering
    \begin{minipage}{\textwidth}
        \centering
        \includegraphics[width=\linewidth]{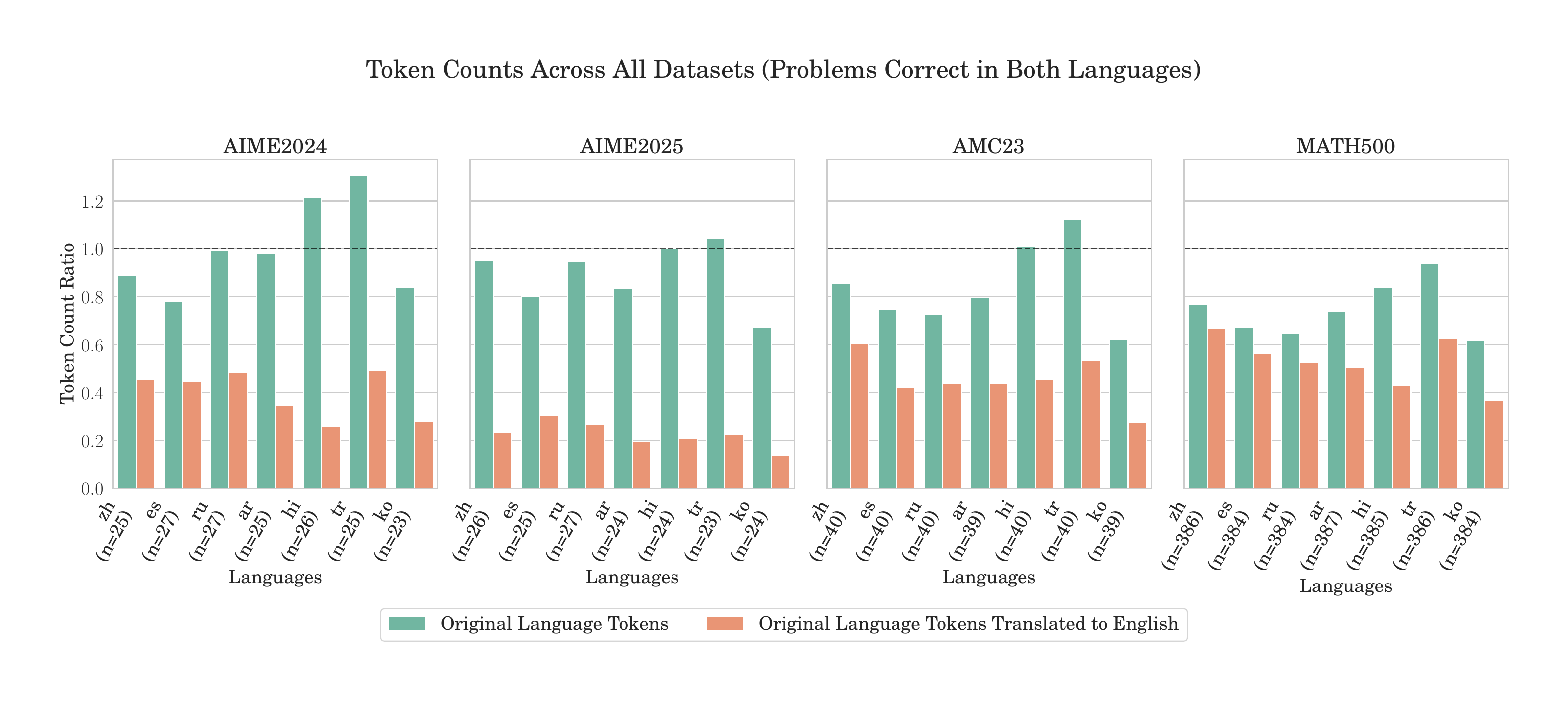}
        \caption{Token Count Ratio vs. number of problems with at least one correct answer in both English and the target language (min. 5 shared cases), computed relative to the average English token count per problem for \textsc{Deepseek R1}.}
        \label{fig:deepseek_token}
    \end{minipage}
    
    \vspace{1em} 
    
    \begin{minipage}{\textwidth}
        \centering
        \renewcommand{\arraystretch}{0.75}
        \setlength{\tabcolsep}{1.5pt}
        \begin{tabular}{c|ccc|ccc|ccc|ccc}
        \toprule
        & \multicolumn{3}{c|}{\textbf{AIME2024}} & \multicolumn{3}{c|}{\textbf{AIME2025}} & \multicolumn{3}{c|}{\textbf{AMC23}} & \multicolumn{3}{c}{\textbf{MATH500}} \\
        \midrule
        \textbf{Lang} & \textbf{N} & \textbf{Win\%} & \textbf{Sig?} & \textbf{N} & \textbf{Win\%} & \textbf{Sig?} & \textbf{N} & \textbf{Win\%} & \textbf{Sig?} & \textbf{N} & \textbf{Win\%} & \textbf{Sig?} \\
        \midrule
        ar & 5 & 100.0 & \textcolor{green!60!black}{\checkmark} & 16 & 93.8 & \textcolor{green!60!black}{\checkmark} & 21 & 100.0 & \textcolor{green!60!black}{\checkmark} & 250 & 92.8 & \textcolor{green!60!black}{\checkmark} \\
        zh & 5 & 100.0 & \textcolor{green!60!black}{\checkmark} & 17 & 94.1 & \textcolor{green!60!black}{\checkmark} & 12 & 58.3 & \textcolor{red}{\ding{55}} & 192 & 55.7 & \textcolor{red}{\ding{55}} \\
        hi & 11 & 100.0 & \textcolor{green!60!black}{\checkmark} & 17 & 88.2 & \textcolor{green!60!black}{\checkmark} & 16 & 93.8 & \textcolor{green!60!black}{\checkmark} & 263 & 95.4 & \textcolor{green!60!black}{\checkmark} \\
        ko & 12 & 100.0 & \textcolor{green!60!black}{\checkmark} & 21 & 100.0 & \textcolor{green!60!black}{\checkmark} & 27 & 96.3 & \textcolor{green!60!black}{\checkmark} & 273 & 90.5 & \textcolor{green!60!black}{\checkmark} \\
        ru & 6 & 83.3 & \textcolor{red}{\ding{55}} & 17 & 82.4 & \textcolor{green!60!black}{\checkmark} & 21 & 90.5 & \textcolor{green!60!black}{\checkmark} & 243 & 93.8 & \textcolor{green!60!black}{\checkmark} \\
        es & 6 & 100.0 & \textcolor{green!60!black}{\checkmark} & 12 & 91.7 & \textcolor{green!60!black}{\checkmark} & 22 & 90.9 & \textcolor{green!60!black}{\checkmark} & 241 & 97.1 & \textcolor{green!60!black}{\checkmark} \\
        tr & 8 & 75.0 & \textcolor{red}{\ding{55}} & 16 & 93.8 & \textcolor{green!60!black}{\checkmark} & 20 & 85.0 & \textcolor{green!60!black}{\checkmark} & 199 & 85.9 & \textcolor{green!60!black}{\checkmark} \\
        \bottomrule
        \end{tabular}
        \captionof{table}{Binomial test results comparing non-English vs English reasoning responses from \textsc{Deepseek R1}. \textcolor{green!60!black}{\checkmark} indicates $p < 0.05$, \textcolor{red}{\ding{55}} indicates no significant difference. $N$ denotes the number of samples wherein both the answers in English and the target language are correct.}
        \label{tab:combined-tables}
    \end{minipage}
    
    \label{fig:deepseek_combined}
    \vspace*{-1.5em}
\end{figure}

To analyze task and language effects in more detail, we examine \textsc{Deepseek R1} (due to its strong TLC) across datasets (Tables~\ref{tab:tlp-delta_deepseek}, \ref{tab:tla_deepseek}, Fig. ~\ref{fig:deepseek_token},~\ref{fig:deepseek_r1_tla}). Detailed results for \textsc{Qwen 2.5} and \textsc{Qwen 3} are provided in Appendix~\ref{sec: detailed_tla_results}. We observe that

\setlength{\parindent}{5pt}

\textbf{-} For \textit{easy} tasks (e.g., MATH500), language choice has little effect on accuracy, while non-English reasoning uses fewer tokens.

\textbf{-} As task difficulty $\uparrow$, we see a relative drop of up to 12. 52\% on average in AIME25 compared to English.

\textbf{-} For harder tasks, high-resource languages (Zh, Ru) outperform low-resource ones (Hi).

\textbf{-} Multilingual reasoning consistently achieves significant token savings across all settings.

\setlength{\parindent}{0pt}

In conclusion, while English remains the top-performing language for reasoning in most LRMs, it isn't always optimal - especially in terms of token efficiency. Models with strong multilingual capabilities can reason effectively in other languages with minimal accuracy loss and significant token savings. For high-resource languages, especially, multilingual reasoning offers a competitive and efficient alternative to English.

\textbf{RQ2: Do multilingual reasoning traces differ meaningfully from their English counterparts?} To answer this, we examine whether token efficiency gains arise from linguistic conciseness or deeper semantic differences in reasoning. We use GPT-4o (temperature = 0.6, top-p = 0.95) to translate non-English traces into English and compare their token counts to the original English versions.

As shown in Table~\ref{fig:deepseek_token}, translated traces remain significantly shorter, indicating that non-English prompts can elicit more concise reasoning paths. This suggests that multilingual prompting can uncover efficient strategies not naturally produced in English. 

In conclusion, multilingual reasoning traces are meaningfully different and often more concise. Their brevity persists after translation, suggesting more efficient reasoning strategies—not just linguistic differences.

\textbf{RQ3: Are Multilingual Reasoning Traces Better Than English?} To further assess whether non-English reasoning traces are \textit{better overall} than their English counterparts, we conduct an automated pairwise comparison using GPT-4o as the judge. For each language, we select cases where both English and non-English reasoning are correct. We then compare the English trace to the GPT-4o-translated non-English trace, leveraging GPT-4o’s high translation quality~\cite{openai2024gpt4ocard} to control for linguistic brevity. Evaluations use GPT-4o with greedy decoding and a prompt (Appendix~\ref{sec: prompts}) focused on clarity, logic, and efficiency. To reduce position bias, we assess each pair in both orders, (A, B) and (B, A) and retain only consistent preferences.
We leverage a one-sided binomial test to establish if reasoning in the native language is (statistically) significantly better than reasoning in English. Table ~\ref{tab:combined-tables} reports results for \textsc{DeepSeek R1}. Across benchmarks, most languages show significant gains when reasoning in the native language, with Arabic, Korean, and Spanish exceeding 90\% win rates. \textsc{Qwen 3} shows similar gains across all non-English languages (Table~\ref{tab:binomial-qwen3}, Appendix~\ref{sec:comparison_results}). We also observe qualitative difference between English outputs and language translated outputs (Ref Appendix \ref{sec:qwen3_example} for a \textsc{Qwen 3} example). We omit \textsc{Qwen 2.5} due to its low Target Language Consistency, which makes such comparisons unreliable (Table~\ref{tab:language-performance-relative}).

These findings suggest that multilingual reasoning can enhance the quality of a model’s thought process. Since all responses are translated to English for evaluation, structural differences are controlled for, indicating that gains likely arise from improved clarity, structure, and conciseness.

\section{Related Work}
\label{sec: related_work}
\textbf{Efficient Reasoning}: Recent interest in LRMs has spurred investigations into reasoning efficiency as to why some tasks demand more reasoning tokens and whether models can answer correctly without explicit reasoning steps \citep{cuadron2025dangeroverthinkingexaminingreasoningaction, hashemi2025dnrbenchbenchmarkingoverreasoning, sui2025stopoverthinkingsurveyefficient, chen2025think23overthinkingo1like, ma2025reasoningmodelseffectivethinking, marjanović2025deepseekr1thoughtologyletsthink}. This work explores reasoning efficiency from a multilingual perspective, analyzing whether certain reasoning paths across languages are both accurate and token-efficient.

\textbf{Multilingual Reasoning}: Prior studies showed that prompting models in the target language often underperforms compared to reasoning in English \cite{qin-etal-2023-cross, shi2022languagemodelsmultilingualchainofthought, ko2025understandsolvetranslatebridging}. However, RLMs like Deepseek R1 and Qwen have shown stronger multilingual reasoning without explicit tuning \cite{deepseekai2025deepseekr1incentivizingreasoningcapability, qwen3, qwen2.5}. We build on this by introducing tighter evaluation bounds and fine-grained trace-level analysis, uncovering language-specific reasoning paths that are both accurate and token-efficient.

Concurrently, several recent works have also begun to explore multilingual reasoning in depth \cite{gao2025thinkingmultilinguallyempowerllm, bajpai2025multilingualtesttimescalinginitial}.
\section{Conclusion and Future Work}
In this work, we explore two central questions: (1) Is English the optimal language for reasoning? and (2) Are multilingual reasoning traces meaningfully different from their English counterparts?

We find that multilingual prompting can reduce token usage by 20–40\% without compromising accuracy, presenting a simple yet effective strategy for improving inference efficiency without retraining. These efficiency gains persist even after translating the outputs back to English, suggesting that the improvements stem from genuine differences in reasoning behavior rather than mere linguistic compactness.

Importantly, the effectiveness of this approach is tied to the model’s underlying multilingual strength. Our findings highlight the need to improve base multilingual alignment to better support efficient and high-quality reasoning across languages. Future work should explore training and preference optimization across more diverse languages to unlock these benefits more broadly.




\section*{Limitations}
In this work, we focus exclusively on reasoning traces generated through English prompts. The behavior of models may vary when prompted in target languages or when guided using techniques like teacher-forcing with a few target-language tokens. Our analysis is limited to seven languages and relies solely on open-source reasoning models. Proprietary models such as OpenAI’s o1 \cite{openai2024openaio1card} may exhibit different behavior from those studied here.

\bibliography{custom}
\bibliographystyle{colm2024_conference}

\appendix
\section{Appendix}
\label{sec:appendix}

\subsection{Prompts}
\definecolor{thinking_color}{RGB}{194, 213, 247}\begin{tcolorbox}[
  title=\textbf{Model Prompt},
  breakable,   
  fonttitle=\bfseries,
  enhanced,                        
  colback=thinking_color!10,           
  colbacktitle=thinking_color,         
  coltitle=black,                 
  colframe=black,    
  coltext=black,                  
  boxrule=0.5pt,
  arc=2mm
]
You \textbf{MUST} give your reasoning AND start in \texttt{\textbf{\{lang\}}} between \texttt{<think>} and \texttt{</think>} tags only. Keep the tags for the thinking part in your response. Do not generate anything outside these tags except for the final answer, which you \textbf{MUST} give in \textbf{ENGLISH}. Please reason step by step in \texttt{\textbf{\{lang\}}} and put your final answer within \texttt{\\boxed\{\{\}\}} in \textbf{ENGLISH}.

\end{tcolorbox}

\vspace{0.5cm}
\begin{figure*}
\centering
\scalebox{0.85}{ 
\begin{tcolorbox}[
  title=\textbf{Evaluation Prompt},
  breakable,
  enhanced,
  fonttitle=\bfseries,
  colback=thinking_color!10,
  colbacktitle=thinking_color,
  coltitle=black,
  colframe=black,
  coltext=black,
  boxrule=0.5pt,
  arc=2mm,
  width=\linewidth,
  floatplacement=H, 
]
\label{appendix:evaluation-prompt}

\textbf{Evaluation Criteria for a Better Response:} \\
You will be comparing two responses that \textbf{both arrive at the correct final answer}. Your task is to evaluate the \textit{quality of the reasoning and explanation} leading to that answer. A better response generally excels in the following areas:

\begin{enumerate}
    \item \textbf{Clarity \& Logic:}
    \begin{itemize}
        \item Presents reasoning in a clear, step-by-step, and logical manner.
        \item Is easy to follow and understand.
        \item Avoids ambiguity in its explanations.
    \end{itemize}
    
    \item \textbf{Efficiency \& Conciseness (Relative to Clarity):}
    \begin{itemize}
        \item Avoids unnecessary detours, excessive repetition, or significant over-explanation of simple concepts \textit{that detract from clarity or add significant bloat}.
    \end{itemize}

    \item \textbf{Structure \& Presentation:}
    \begin{itemize}
        \item Organizes the solution in a well-structured way.
        \item May include helpful checks or clarifications that genuinely aid understanding without being redundant or overly verbose.
    \end{itemize}
\end{enumerate}

\vspace{1em}
\noindent
\textbf{Important Note on Positional Bias:} The order in which responses are presented below (A then B) is arbitrary. Please strive to evaluate each response on its own merits against the criteria before making a comparative judgment. Your final decision should not be influenced by this presentation order.

\vspace{1.5em}
\textbf{Your Task:} \\
You are shown a math question and two responses, labeled ``A'' and ``B''. Please follow these steps in your evaluation process:

\begin{enumerate}
    \item \textbf{Assess Individually:}
    \begin{itemize}
        \item First, carefully read and assess the response labeled ``A'' based \textit{solely} on the ``Evaluation Criteria'' above. Note its strengths and weaknesses.
        \item Next, carefully read and assess the response labeled ``B'' based \textit{solely} on the ``Evaluation Criteria'' above. Note its strengths and weaknesses.
    \end{itemize}

    \item \textbf{Compare and Decide:}
    \begin{itemize}
        \item Now, compare your independent assessments of ``A'' and ``B''.
        \item Decide \textbf{which response is better overall}.
        \item If they are of very similar quality, or if their different strengths and weaknesses balance out making a clear preference difficult, you may indicate they are \textit{``Neither -- Roughly Equal''}.
    \end{itemize}

    \item \textbf{Explain Your Decision:}
    \begin{itemize}
        \item Provide a \textbf{brief but specific explanation} for your final comparative choice. Justify why the chosen response is superior, or why they are roughly equal, referencing specific aspects of the criteria and your comparative judgment. \\
        \textit{Example:} ``A was ultimately preferred because its clearer logical flow outweighed its slight verbosity compared to B.'' \\
        \textit{Example:} ``Neither -- Roughly Equal because while A was more concise, B offered a more intuitive structural breakdown needed for this question.''
    \end{itemize}
\end{enumerate}

\vspace{1.5em}
\textbf{Question:} \texttt{\{question\}} \\
\textbf{Response A:} \texttt{\{response\_1\_text\}} \\
\textbf{Response B:} \texttt{\{response\_2\_text\}}

\vspace{1em}
\noindent\textbf{Respond ONLY in this JSON format:}

\vspace{0.5em}

\begin{minipage}{\linewidth}
\begin{ttfamily}
\noindent\{\\
\quad "better\_response\_label": "A" or "B" or "Neither - Roughly Equal",\\
\quad "explanation": "<Your specific explanation for the final comparative decision, reflecting the comparison>"\\
\}
\end{ttfamily}
\end{minipage}

\end{tcolorbox}
}
\end{figure*}
\label{sec: prompts}

\subsection{Unbiased Estimator for \texttt{TLP@k}}
\label{app:unbiased_estimator}

For a given language $l$ and $k$, we want to estimate 
\begin{equation}
    Pass@k(l) = \mathbb{E}_{x}\left[\mathbbm{1}(\exists r \in \mathcal{R}^{(k)}(x) | LLM(x, r) = y \wedge LID(r) = l\right]
\end{equation}
The naive estimator for the above on a dataset $\mathcal{D} = \{(x_0, y_0), ... (x_m, y_m)\}$ can be defined as
\begin{equation}
\hat{Pass@k(l)} = \frac{1}{m} \sum_{i=1}^{m}\left[\mathbbm{1}(\exists r \in \mathcal{R}^{(k)}(x_{i}) | LLM(x_{i}, r) = y_{i} \wedge LID(r) = l\right]
\end{equation}
However, as shown in \cite{chen2021evaluatinglargelanguagemodels}, this estimator has a high variance. Consequently, we extend the low variance estimator proposed in \cite{chen2021evaluatinglargelanguagemodels} for estimating $Pass@k(l)$ as

\begin{equation}
\begin{aligned}
\hat{Pass@k(l, n)} &= \frac{1}{m} \sum_{i=1}^{m}\left[1 - \frac{\binom{n - c\left(x_i, y_i\right)}{k}}{\binom{n}{k}}\right] \\
c(x_i, y_i) &= \sum_{r \in R^{(n)}(x_i)} \mathbbm{1}\left( LLM(x_{i}, r) = y_i \wedge LID(r) = l\right)
\end{aligned}
\end{equation}

Thus for a fixed $k$ by increasing the number of reasoning traces $n$, we reduce the variance of the estimator. Calculating this estimator directly results in very large numbers and numerical instability. Instead, we use \ref{lst:pass_at_k} below, as done in  \cite{chen2021evaluatinglargelanguagemodels}.

\vspace{0.5cm}
\begin{lstlisting}[language=Python, basicstyle=\footnotesize\ttfamily, label={lst:pass_at_k}]
def pass_at_k(n, c, k):
    """
    :param n: total number of samples generated
    :param c: number of correct samples as defined above
    :param k: k in TLP@k%
    """
    if n - c < k: return 1.0
    return 1.0 - np.prod(1.0 - k /
        np.arange(n - c + 1, n + 1))
\end{lstlisting}

\subsection{\texttt{TLA} Results}
To provide a finer-grained view, we also define Target Language Accuracy (TLA) as the average accuracy over faithful traces:

\begin{equation}
\texttt{TLA}(j) = \frac{1}{m} \sum_{i=1}^{m} \frac{\sum_{r \in \mathcal{T}(i,j)} \mathbbm{1}\left[ \texttt{LLM}(r) = y_i \right]}{|\mathcal{T}(i,j)|}
\end{equation}
\label{eq::acc}

\begin{figure*}
    \centering
        \includegraphics[width=0.8\textwidth]{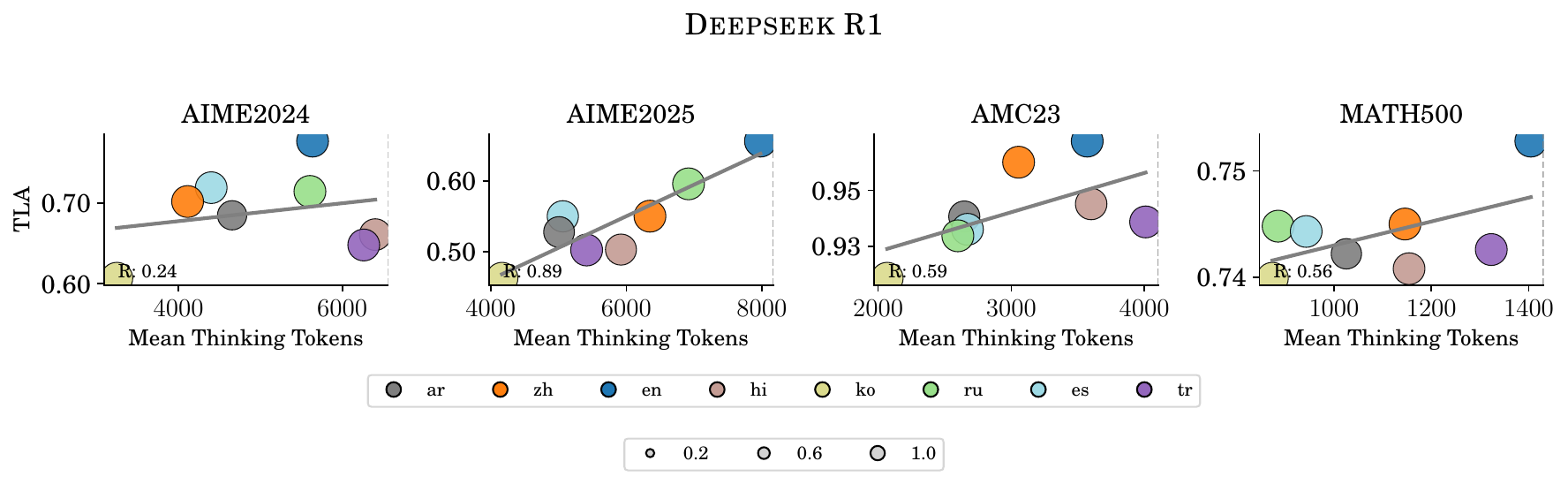}
        \caption{\textsc{Deepseek R1} analysis: \texttt{TLA} versus mean thinking tokens per language, with point size indicating how often the target language was used (out of 16 samples).}
        \label{fig:deepseek_r1_tla}
    
    
\end{figure*}

\begin{figure*}
    \centering
    \begin{subfigure}[b]{0.8\textwidth}
        \centering
        \includegraphics[width=\linewidth]{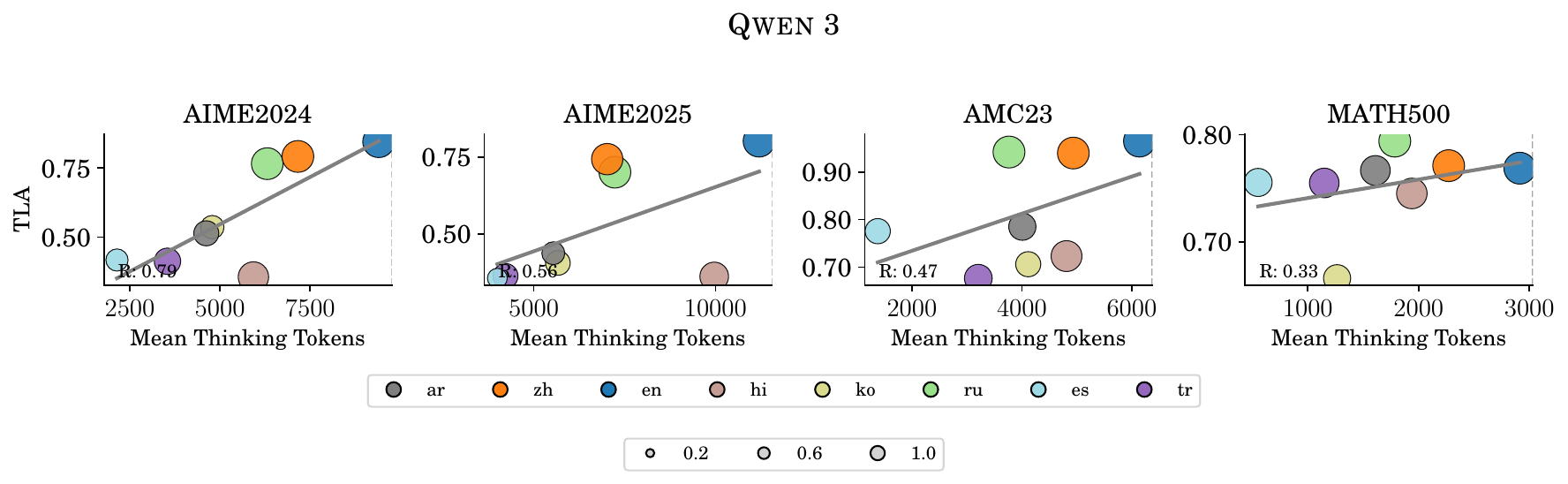}
        \caption{Figure~\ref{fig:qwen3_tla} shows \texttt{TLA} versus mean thinking tokens per language, with point size indicating how often the target language was used (out of 16 samples).}
        \label{fig:qwen3_tla}
    \end{subfigure}
    
    \vspace{1em} 
    
    \begin{subfigure}[b]{0.8\textwidth}
        \centering
        \includegraphics[width=\linewidth]{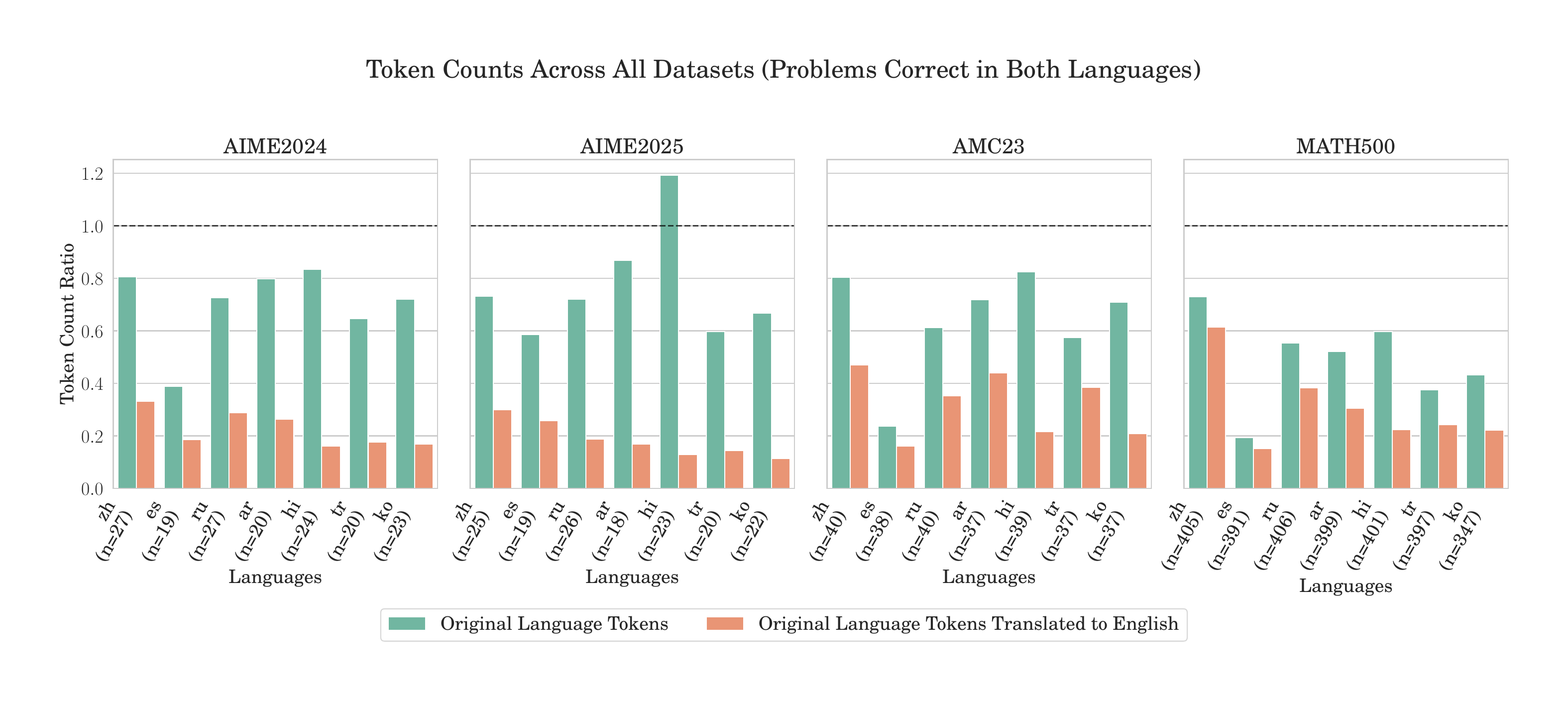}
        \caption{Figure~\ref{fig:qwen3_token} plots the Token Count Ratio against the number of problems with at least one correct answer in both English and the target language (minimum 5 shared instances). The ratio is computed relative to the average English token count for each problem.}
        \label{fig:qwen3_token}
    \end{subfigure}
    \caption{\textsc{Qwen 3} analysis}
    \label{fig:qwen3_combined}
\end{figure*}

\begin{figure*}
    \centering
    \begin{subfigure}[b]{0.8\textwidth}
        \centering
        \includegraphics[width=\linewidth]{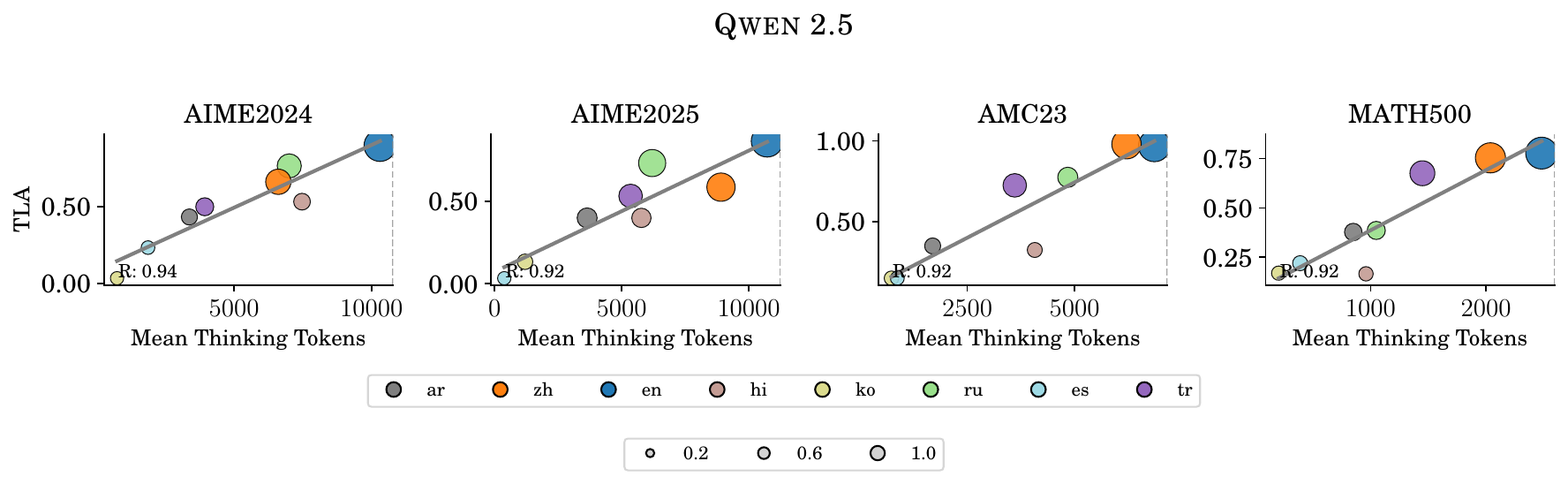}
        \caption{Figure~\ref{fig:qwen25_tla} shows \texttt{TLA} versus mean thinking tokens per language, with point size indicating how often the target language was used (out of 16 samples).}
        \label{fig:qwen25_tla}
    \end{subfigure}
    
    \vspace{1em} 
    
    \begin{subfigure}[b]{0.8\textwidth}
        \centering
        \includegraphics[width=\linewidth]{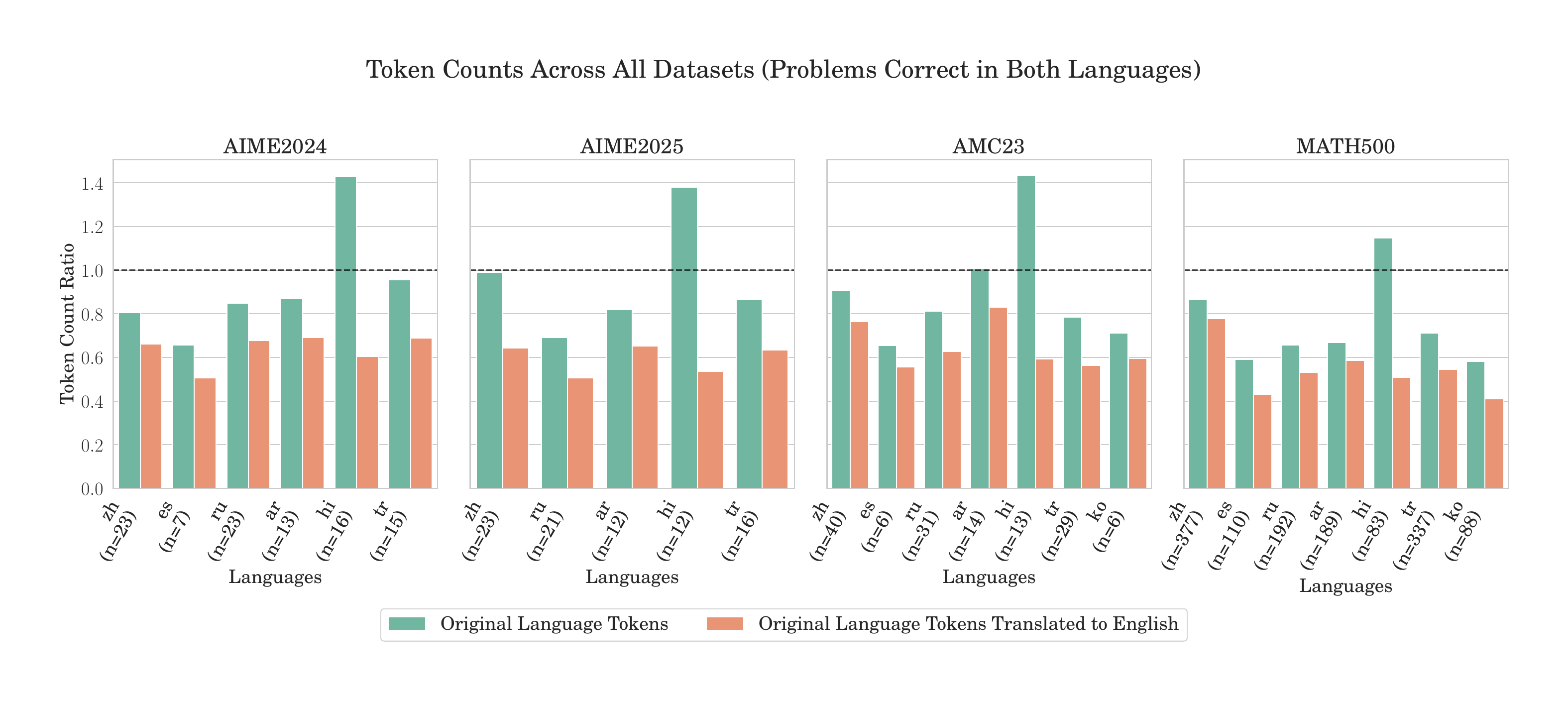}
        \caption{Figure~\ref{fig:qwen25_token} plots the Token Count Ratio against the number of problems with at least one correct answer in both English and the target language (minimum 5 shared instances). The ratio is computed relative to the average English token count for each problem.}
        \label{fig:qwen25_token}
    \end{subfigure}
    \caption{\textsc{Qwen 2.5} analysis}
    \label{fig:qwen25_combined}
\end{figure*}

Detailed \texttt{TLA} results for \textsc{Deepseek R1} in Table \ref{tab:tla_deepseek}  and Figure \ref{fig:deepseek_r1_tla}, for \textsc{Qwen 2.5} in Table \ref{tab:tla_qwen2_5} and Figure \ref{fig:qwen25_combined}. For \textsc{Qwen 3} in Table \ref{tab:tla_qwen3} and Figure \ref{fig:qwen3_combined}.

\label{sec: detailed_tla_results}

\begin{table*}[htb]
  \centering
  \begin{tabular}{lrrrrrrrr}
    \toprule
    \textbf{Dataset} & \textbf{en} & \textbf{zh} & \textbf{ru} & \textbf{es} & \textbf{ar} & \textbf{tr} & \textbf{hi} & \textbf{ko} \\
    \midrule
    MATH500  & 0.75 & 0.75 & 0.74 & 0.74 & 0.74 & 0.74 & 0.74 & 0.74 \\
    AMC23    & 0.97 & 0.96 & 0.93 & 0.93 & 0.94 & 0.94 & 0.94 & 0.91 \\
    AIME2024 & 0.78 & 0.70 & 0.71 & 0.72 & 0.69 & 0.65 & 0.66 & 0.61 \\
    AIME2025 & 0.66 & 0.55 & 0.60 & 0.55 & 0.53 & 0.50 & 0.50 & 0.46 \\
    \bottomrule
  \end{tabular}
  \caption{\texttt{TLA} scores as defined in Equation: \ref{eq::acc} for \textsc{Deepseek R1}}
  \label{tab:tla_deepseek}
\end{table*}
\begin{table*}[htb]
  \centering
  
  \begin{tabular}{lrrrrrrrr}
    \toprule
    \textbf{Dataset} & \textbf{en} & \textbf{zh} & \textbf{ru} & \textbf{es} & \textbf{ar} & \textbf{tr} & \textbf{hi} & \textbf{ko} \\
    \midrule
    MATH500  & 0.77 & 0.77 & 0.79 & 0.76 & 0.77 & 0.76 & 0.75 & 0.67 \\
    AMC23    & 0.97 & 0.94 & 0.94 & 0.78 & 0.79 & 0.68 & 0.72 & 0.71 \\
    AIME2024 & 0.85 & 0.79 & 0.77 & 0.42 & 0.51 & 0.41 & 0.36 & 0.54 \\
    AIME2025 & 0.80 & 0.74 & 0.70 & 0.36 & 0.44 & 0.36 & 0.36 & 0.41 \\
    \bottomrule
  \end{tabular}
  \caption{\texttt{TLA} scores as defined in Equation: \ref{eq::acc} for \textsc{Qwen 2.5}}
  \label{tab:tla_qwen2_5}
\end{table*}
\begin{table*}[htb]
  \centering
  
  \begin{tabular}{lrrrrrrrr}
    \toprule
    \textbf{Dataset} & \textbf{en} & \textbf{zh} & \textbf{ru} & \textbf{es} & \textbf{ar} & \textbf{tr} & \textbf{hi} & \textbf{ko} \\
    \midrule
    MATH500  & 0.77 & 0.77 & 0.79 & 0.76 & 0.77 & 0.76 & 0.75 & 0.67 \\
    AMC23    & 0.97 & 0.94 & 0.94 & 0.78 & 0.79 & 0.68 & 0.72 & 0.71 \\
    AIME2024 & 0.85 & 0.79 & 0.77 & 0.42 & 0.51 & 0.41 & 0.36 & 0.54 \\
    AIME2025 & 0.80 & 0.74 & 0.70 & 0.36 & 0.44 & 0.36 & 0.36 & 0.41 \\
    \bottomrule
  \end{tabular}
  \caption{\texttt{TLA} scores as defined in Equation: \ref{eq::acc} for \textsc{Qwen 3}}
  \label{tab:tla_qwen3}
\end{table*}
\definecolor{green-1}{RGB}{213, 216, 123}
\definecolor{green-2}{RGB}{184, 209, 143}
\definecolor{green-3}{RGB}{152, 199, 134}
\definecolor{green-4}{RGB}{101, 189, 110}
\definecolor{yellow-1}{RGB}{255, 234, 132}
\definecolor{yellow-2}{RGB}{254, 217, 129}
\definecolor{orange-1}{RGB}{254, 206, 127}
\definecolor{orange-2}{RGB}{252, 173, 120}
\definecolor{red-1}{RGB}{247, 121, 121}


\begin{table*}[t]
\centering

\begin{subtable}{\textwidth}
\centering
\resizebox{\textwidth}{!}{%
\begin{tabular}{l|cc|cc|cc|cc|cc|cc|cc|cc}
\toprule
& \multicolumn{2}{c|}{\textbf{en}} & \multicolumn{2}{c|}{\textbf{zh}} & \multicolumn{2}{c|}{\textbf{ru}} & \multicolumn{2}{c|}{\textbf{es}} & \multicolumn{2}{c|}{\textbf{ar}} & \multicolumn{2}{c|}{\textbf{tr}} & \multicolumn{2}{c|}{\textbf{hi}} & \multicolumn{2}{c}{\textbf{ko}} \\
\cmidrule{2-17}
\textbf{Dataset} & \textbf{TLP@k} & \textbf{Tokens}
& $\Delta$\%TLP & $\Delta$\%Tok
& $\Delta$\%TLP & $\Delta$\%Tok
& $\Delta$\%TLP & $\Delta$\%Tok
& $\Delta$\%TLP & $\Delta$\%Tok
& $\Delta$\%TLP & $\Delta$\%Tok
& $\Delta$\%TLP & $\Delta$\%Tok
& $\Delta$\%TLP & $\Delta$\%Tok \\
\midrule
\textbf{MATH500} & 0.78 & 2482.11 & 78.21 & 61.31 & 14.10 & 41.62 & 3.85 & 17.86 & 71.79 & 84.31 & 51.28 & 65.76 & 51.28 & 57.73 & 76.92 & 91.80 \\
\textbf{AMC23} & 1.00 & 6860.88 & 67.00 & 40.64 & 27.00 & 47.45 & 0.00 & 9.48 & 85.00 & 87.35 & 65.00 & 75.34 & 22.00 & 29.43 & 85.00 & 89.34 \\
\textbf{AIME2024} & 0.90 & 10310.63 & 41.11 & 27.53 & 44.44 & 61.88 & 14.44 & 35.90 & 74.44 & 81.92 & 52.22 & 67.30 & 14.44 & 32.04 & 96.67 & 92.85 \\
\textbf{AIME2025} & 0.87 & 10725.53 & 54.02 & 46.16 & 39.08 & 50.12 & 11.49 & 16.95 & 96.55 & 96.49 & 54.02 & 66.08 & 19.54 & 42.21 & 85.06 & 88.86 \\
\bottomrule
\end{tabular}%
}
\caption{\textsc{Qwen 2.5}}
\end{subtable}
\begin{subtable}{\textwidth}
\centering
\resizebox{\textwidth}{!}{%
\begin{tabular}{l|cc|cc|cc|cc|cc|cc|cc|cc}
\toprule
& \multicolumn{2}{c|}{\textbf{en}} & \multicolumn{2}{c|}{\textbf{zh}} & \multicolumn{2}{c|}{\textbf{ru}} & \multicolumn{2}{c|}{\textbf{es}} & \multicolumn{2}{c|}{\textbf{ar}} & \multicolumn{2}{c|}{\textbf{tr}} & \multicolumn{2}{c|}{\textbf{hi}} & \multicolumn{2}{c}{\textbf{ko}} \\
\cmidrule{2-17}
\textbf{Dataset} & \textbf{TLP@k} & \textbf{Tokens}
& $\Delta$\%TLP & $\Delta$\%Tok
& $\Delta$\%TLP & $\Delta$\%Tok
& $\Delta$\%TLP & $\Delta$\%Tok
& $\Delta$\%TLP & $\Delta$\%Tok
& $\Delta$\%TLP & $\Delta$\%Tok
& $\Delta$\%TLP & $\Delta$\%Tok
& $\Delta$\%TLP & $\Delta$\%Tok \\
\midrule
\textbf{MATH500} & 0.81 & 2913.68 & 0.00 & 22.05 & 0.00 & 38.74 & 3.70 & 81.08 & 1.23 & 44.75 & 2.47 & 60.59 & 1.23 & 33.46 & 14.81 & 56.61 \\
\textbf{AMC23} & 1.00 & 6138.45 & 0.00 & 19.60 & 0.00 & 38.67 & 5.00 & 77.58 & 7.00 & 34.71 & 7.00 & 47.76 & 2.00 & 21.61 & 7.00 & 32.97 \\
\textbf{AIME2024} & 0.93 & 9399.33 & 3.23 & 23.83 & 3.23 & 32.86 & 32.26 & 77.18 & 27.96 & 50.84 & 27.96 & 62.29 & 13.98 & 36.94 & 17.20 & 49.07 \\
\textbf{AIME2025} & 0.93 & 11200.1 & 10.75 & 37.32 & 6.45 & 35.43 & 32.26 & 64.32 & 35.48 & 50.57 & 27.96 & 62.32 & 17.20 & 11.02 & 21.51 & 49.44 \\
\bottomrule
\end{tabular}%
}
\caption{\textsc{Qwen 3}}
\end{subtable}

\vspace{0.5em}

\caption{Performance and token efficiency of native-language reasoning using \textsc{Qwen} models. In each subtable, \texttt{TLP@4} and mean token count are reported as percentage change from English}
\label{tab:tla-delta-qwen-stacked}
\end{table*}
\begin{table*}[th]
\centering
\small

\subfloat[Performance across English, Chinese, Russian, and Spanish.]{
\begin{tabular}{l|ccc|ccc|ccc|ccc}
\toprule
\textbf{Model} & \multicolumn{3}{c|}{\textbf{en}} & \multicolumn{3}{c|}{\textbf{zh}} & \multicolumn{3}{c|}{\textbf{ru}} & \multicolumn{3}{c}{\textbf{es}} \\
\cmidrule(lr){2-4} \cmidrule(lr){5-7} \cmidrule(lr){8-10} \cmidrule(lr){11-13}
& \texttt{TLP@4} & Con. & Tok. & \texttt{TLP@4} & Con. & Tok. & \texttt{TLP@4} & Con. & Tok. & \texttt{TLP@4} & Con. & Tok. \\
\midrule
\textsc{DeepSeek R1} & 0.90 & 1.00 & 3816 & 0.87 & 1.00 & 3009 & 0.89 & 1.00 & 3276 & 0.88 & 0.99 & 2677 \\
\textsc{Qwen 2.5}     & 0.89 & 1.00 & 6249 & 0.82 & 0.80 & 4912 & 0.66 & 0.34 & 3861 & 0.16 & 0.06 & 746 \\
\textsc{Qwen 3}     & 0.92 & 1.00 & 6123 & 0.89 & 0.99 & 4427 & 0.90 & 1.00 & 3932 & 0.75 & 0.58 & 1656 \\
\bottomrule
\end{tabular}
}

\vspace{1em}

\subfloat[Performance across Arabic, Turkish, Hindi, and Korean.]{
\begin{tabular}{l|ccc|ccc|ccc|ccc}
\toprule
\textbf{Model} & \multicolumn{3}{c|}{\textbf{ar}} & \multicolumn{3}{c|}{\textbf{tr}} & \multicolumn{3}{c|}{\textbf{hi}} & \multicolumn{3}{c}{\textbf{ko}} \\
\cmidrule(lr){2-4} \cmidrule(lr){5-7} \cmidrule(lr){8-10} \cmidrule(lr){11-13} 
& \texttt{TLP@4} & Con. & Tok. & \texttt{TLP@4} & Con. & Tok. & \texttt{TLP@4} & Con. & Tok. & \texttt{TLP@4} & Con. & Tok. \\
\midrule
\textsc{DeepSeek R1} & 0.85 & 0.90 & 2738 & 0.84 & 1.00 & 3498 & 0.86 & 0.97 & 3521 & 0.83 & 1.00 & 2140 \\
\textsc{Qwen 2.5}     & 0.39 & 0.18 & 1996 & 0.61 & 0.44 & 2978 & 0.36 & 0.10 & 3668 & 0.12 & 0.06 & 616 \\
\textsc{Qwen 3}     & 0.75 & 0.67 & 3223 & 0.77 & 0.72 & 2476 & 0.84 & 0.89 & 4675 & 0.78 & 0.57 & 3221 \\
\bottomrule
\end{tabular}
}

\caption{Performance metrics (\texttt{TLP@4}, Consistency, Mean Token count) across different languages and models.}
\label{tab:language-performance}
\end{table*}

\subsection{Detailed Comparison Test Results}
Check Table~\ref{tab:binomial-qwen3} for the results for \textsc{Qwen 3}.
\label{sec:comparison_results}
\begin{table*}[t]
\centering
\small
\setlength{\tabcolsep}{3pt}
\begin{tabular}{c|ccc|ccc|ccc|ccc}
\toprule
& \multicolumn{3}{c|}{\textbf{AIME2024}} & \multicolumn{3}{c|}{\textbf{AIME2025}} & \multicolumn{3}{c|}{\textbf{AMC23}} & \multicolumn{3}{c}{\textbf{MATH500}} \\
\midrule
Lang & N & Win (\%) & Sig? & N & Win (\%) & Sig? & N & Win (\%) & Sig? & N & Win (\%) & Sig? \\
\midrule
ar & 12 & 100.0 & \textcolor{green!60!black}{\checkmark} & 10 & 100.0 & \textcolor{green!60!black}{\checkmark} & 21 & 100.0 & \textcolor{green!60!black}{\checkmark} & 295 & 98.6 & \textcolor{green!60!black}{\checkmark} \\
zh & 11 & 90.9 & \textcolor{green!60!black}{\checkmark} & 13 & 100.0 & \textcolor{green!60!black}{\checkmark} & 11 & 81.8 & \textcolor{green!60!black}{\checkmark} & 94 & 73.4 & \textcolor{green!60!black}{\checkmark} \\
hi & 20 & 100.0 & \textcolor{green!60!black}{\checkmark} & 18 & 94.4 & \textcolor{green!60!black}{\checkmark} & 34 & 94.1 & \textcolor{green!60!black}{\checkmark} & 368 & 98.1 & \textcolor{green!60!black}{\checkmark} \\
ko & 17 & 100.0 & \textcolor{green!60!black}{\checkmark} & 19 & 94.7 & \textcolor{green!60!black}{\checkmark} & 31 & 96.8 & \textcolor{green!60!black}{\checkmark} & 298 & 99.3 & \textcolor{green!60!black}{\checkmark} \\
ru & 12 & 91.7 & \textcolor{green!60!black}{\checkmark} & 15 & 100.0 & \textcolor{green!60!black}{\checkmark} & 21 & 95.2 & \textcolor{green!60!black}{\checkmark} & 252 & 99.6 & \textcolor{green!60!black}{\checkmark} \\
es & 16 & 93.8 & \textcolor{green!60!black}{\checkmark} & 12 & 91.7 & \textcolor{green!60!black}{\checkmark} & 36 & 100.0 & \textcolor{green!60!black}{\checkmark} & 370 & 100.0 & \textcolor{green!60!black}{\checkmark} \\
tr & 18 & 94.4 & \textcolor{green!60!black}{\checkmark} & 15 & 93.3 & \textcolor{green!60!black}{\checkmark} & 27 & 100.0 & \textcolor{green!60!black}{\checkmark} & 345 & 99.1 & \textcolor{green!60!black}{\checkmark} \\
\bottomrule
\end{tabular}
\caption{Binomial test results comparing non-English vs English reasoning responses obtained from \textsc{Qwen 3}, using consistent GPT-4o judgments. \textcolor{green!60!black}{\checkmark} indicates $p < 0.05$, and \textcolor{red}{\ding{55}} indicates no significant difference.}
\label{tab:binomial-qwen3}
\end{table*}

\subsection{Qualitative analysis done for a \textsc{Qwen 3} output in Chinese}
\label{sec:qwen3_example}
Check Figure \ref{fig:example_qwen3_chinese_output} and Table \ref{tab:overthinking_comparison} for a qualitative check on \textsc{Qwen 3} output.
\begin{figure*}
\centering
\scalebox{0.85}{
\definecolor{thinking_color}{RGB}{194, 213, 247}\begin{tcolorbox}[
  title=\textbf{Example problem from MATH500 solved by \textsc{Qwen 3} in Chinese},
  breakable,   
  fonttitle=\bfseries,
  enhanced,                        
  colback=thinking_color!10,           
  colbacktitle=thinking_color,         
  coltitle=black,                 
  colframe=black,    
  coltext=black,                  
  boxrule=0.5pt,
  arc=2mm
]

\textbf{Question}: Simplify $\sqrt{242}$. \\

\textbf{Response in English}: Okay, so I need to simplify the square root of $242$. Hmm, let me think. First, I remember that simplifying a square root involves factoring the number into its prime factors and then looking for pairs of the same number because each pair can be taken out of the square root as a single number.

Alright, starting with $242$. Let me try dividing by small primes first. The number $242$ is even, right? Because it ends with a $2$, so it should be divisible by $2$. Let me check: $242$ divided by $2$ equals $121$. So that gives me $2$ times $121$. Now, I need to factor $121$. Hmm, $121$... I think that's a square number. Wait, $11$ times $11$ is $121$. Yes, that's right! So putting that all together, the prime factorization of $242$ is $2 \times 11 \times 11$.

So, writing that out as a product of primes: $2 \times 11^2$. Now, when simplifying square roots, every pair of primes can be taken out of the square root. Here, we have a pair of $11$s and a single $2$. That means the square root of $242$ can be written as $\sqrt{11^2 \times 2}$.

Using the property of square roots that $\sqrt{a \times b} = \sqrt{a} \times \sqrt{b}$, this becomes $\sqrt{11^2} \times \sqrt{2}$. Since $\sqrt{11^2}$ is $11$, the simplified form should be $11 \times \sqrt{2}$.

Wait a second, let me double-check. If I square $11\sqrt{2}$, I should get back the original number. Let's see: $(11\sqrt{2})^2 = 11^2 \times (\sqrt{2})^2 = 121 \times 2 = 242$. Yep, that checks out.

But hold on, did I factorize correctly? Let me confirm. $2 \times 121 = 242$. Then $121$ is $11$ squared. Correct. So there are no other factors. There's a single $2$ and a pair of $11$s. So taking the pair out, leaving the $2$ inside.

Is there a possibility that $242$ could be factored differently? Let's see. Suppose I didn't start with $2$. What if I tried dividing by $11$ first? Let's try: $242$ divided by $11$ is $22$. Then $22$ divided by $11$ is $2$. So that gives the same result: $11 \times 11 \times 2$. So regardless of the order, the prime factors are the same.
Therefore, regardless of the method, the simplified square root should be $11\sqrt{2}$.

I think that's all. I don't see any mistakes in my reasoning. So the answer should be $11$ times the square root of $2$. \\

\textbf{Translated Response for Chinese to English}: First, I need to simplify $\sqrt{242}$. To simplify the square root, I need to find the square factors of the number under the root. First, I will factorize $242$ into its prime factors. Dividing $242$ by $2$ gives $121$, so $242 = 2 \times 121$. Next, $121$ is a known square number because $11 \times 11 = 121$. Therefore, $\sqrt{242}$ can be broken down into $\sqrt{2 \times 121}$.

According to the multiplication property of square roots, $\sqrt{a \times b} = \sqrt{a} \times \sqrt{b}$, so $\sqrt{2 \times 121} = \sqrt{2} \times \sqrt{121}$. Since $\sqrt{121} = 11$, the entire expression can be simplified to $11\sqrt{2}$.

Therefore, the simplified form of $\sqrt{242}$ is $11\sqrt{2}$.

\end{tcolorbox}
}
\caption{Example problem solved by \textsc{Qwen 3}}
\label{fig:example_qwen3_chinese_output}
\end{figure*}
\begin{table}[ht]
\tiny
\setlength{\tabcolsep}{2pt}
\renewcommand{\arraystretch}{1.1}
\centering
\begin{tabular}{p{2.1cm}p{2.1cm}p{2.1cm}}
\toprule
\textbf{Phenomenon} & \textbf{English Response} & \textbf{Chinese Response} \\
\midrule
Self-dialogue & “Hmm, let me think...” & “First, simplify $\sqrt{242}$.” \\
Verification & “11×11 is 121... Yes!” & “121 is a known square number...” \\
Redundant check & “Let me double-check...” & No equivalent step \\
Alternative path & “What if I didn’t start with 2...” & No alternative explored \\
Repeated confirmation & “I don’t see any mistakes...” & “$\sqrt{242}$ is $11\sqrt{2}$.” \\
\bottomrule
\end{tabular}
\caption{Comparison of English and translated Chinese traces (Qwen 3, MATH500). English shows overthinking. Full output in Figure~\ref{fig:example_qwen3_chinese_output}.}
\label{tab:overthinking_comparison}
\end{table}


\end{document}